# Deep Video Generation, Prediction and Completion of Human Action Sequences

Haoye Cai[*]　　Chunyan Bai[*]　　　　Yu-Wing Tai　　　　Chi-Keung Tang
HKUST　　　　　　　　　　　Tencent YouTu　　　　　　HKUST
{hcaiaa,cbai}@connect.ust.hk　　yuwingtai@tencent.com　　cktang@cs.ust.hk

**Abstract**

*Current deep learning results on video generation are limited while there are only a few first results on video prediction and no relevant significant results on video completion. This is due to the severe ill-posedness inherent in these three problems. In this paper, we focus on human action videos, and propose a general, two-stage deep framework to generate human action videos with no constraints or arbitrary number of constraints, which uniformly address the three problems: video generation given no input frames, video prediction given the first few frames, and video completion given the first and last frames[1]. To make the problem tractable, in the first stage we train a deep generative model that generates a human pose sequence from random noise. In the second stage, a skeleton-to-image network is trained, which is used to generate a human action video given the complete human pose sequence generated in the first stage. By introducing the two-stage strategy, we sidestep the original ill-posed problems while producing for the first time high-quality video generation/prediction/completion results of much longer duration. We present quantitative and qualitative evaluation to show that our two-stage approach outperforms state-of-the-art methods in video generation, prediction and video completion. Our video result demonstration can be viewed at https://iamacewhite.github.io/supp/index.html.*

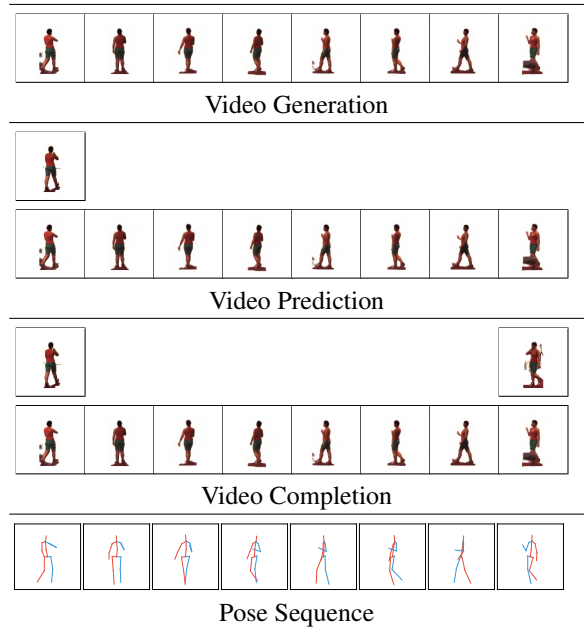

Figure 1. From top: video generation (from scratch), prediction and completion of human action videos using our general two-stage deep framework. In all cases, a complete human pose skeleton sequence is generated in the first stage, shown in bottom.

## 1. Introduction

In this paper we propose a general, two-stage deep framework for human video generation, prediction and completion (Figure 1), where each problem was previously addressed as separate problems. Previous video generation capitalizing state-of-the-art deep convolutional neural network (CNN), such as [35], has demonstrated the significant difficulty of the problem, where their first results were still far from photorealism. Current future prediction [20] in the form of video prediction [37] generates a short video from a given frame to predict future actions in a very limited scope. While there exist deep learning works on image completion [43], there is no known representative deep learning work on video completion.

To better address the general video synthesis problem, we need to understand how pixels change to generate a full temporal object action. With a higher level of uncertainty in the exact movement between frames of the moving object, as observed in [37], the problem is more tractable by modeling the uncertainty with underlying structure of the moving objects, which in our case is human poses. Human action videos are arguably the most interesting and useful videos in various computer vision applications. Thus, we focus on human action videos in this paper, and divide the task into human pose sequence generation (pose space) followed by image generation (pixel space) from the generated human pose sequences.

Specifically, our general deep framework for video generation has two stages: first, a new conditional generative adversarial network (GAN) is proposed to generate a plausible pose sequence that performs a given category of actions; we then apply our supervised reconstruction network with feature matching loss to transfer pose sequence to the pixel space in order to generate the full output video. Our general video generation framework can be specialized to video completion and video prediction by optimizing in the latent space to generate video results that best suit the given input constraints. Hence our approach can either generate videos from scratch, or complete/predict a video with arbitrary number of input frames available given the action class. We will provide extensive qualitative and quantitative experimental results to demonstrate that our model is able to generate and complete natural human motion video under the constraints prescribed by input frames if available.

---

[*]Equal Contribution.
[1]Without causing confusion, we refer unconstrained video generation as just video generation when no input frames are given, and we still use video prediction to refer to input situations where the first few frames are given, i.e., not just the first frame.

## 2. Related Work

Our general framework and implementation can be uniformly applied to video generation, prediction and completion. Video completion or prediction can be regarded as video generation constrained by the input frames. This paper focuses on human action videos, and our two-step approach consists of human pose estimation and image generation. We review here recent representative works leveraging deep learning to achieve state-of-the-art results.

**Video Prediction/Generation** In video forecasting, research has been done to model uncertain human motion in pose space [37]. Attempts have also been made to learn the deep predictive coding [18] or various feature representation [20, 36]. To generate videos from scratch, work has been done to generate videos directly in pixel space [35]. While these works shed some light on how we should model the uncertainty and temporal information in videos, our work aims at a different goal: video completion and generation in the same framework. Video prediction is useful in training computer agents to play computer games or how to act in complicated 2D or 3D environment [20] by predicting the next frames (consisting of simple objects in [20, 18, 8]).

**Image/Video Completion** Much work in the completion area has been focused on image completion with Generative Models [43]. However, video completion in the realm of deep learning has remain unexplored, although video completion is an important topic [14, 41]. If the temporal distance between the input frames is small, e.g., [24] then videos frame interpolation can be performed to fill in the in-between frames. We are dealing with a different problem where input frames are far apart from each other in completion. The modeling of such uncertainty increased the difficulty of this task. In our paper, we aim to perform video completion by optimizing the latent space under the constraint of input frames.

**Human Pose Estimation** Various research efforts have been made to produce state-of-the-art human pose estimation results, providing us with reliable human pose baseline. Realtime multi-person pose estimation was achieved in [3]. Other important works include DeepPose [34], Pose Machine [40] and Adversarial PoseNet [5]. The current state-of-the-art human pose estimation on our choice of dataset, Human3.6m [13], is achieved by Newell et al [23]. In our paper, we leverage the reliable human pose estimation from [23] in human motion videos and complete a pose sequence that looks smooth and natural.

**Generative Models** Our work is based on Generative Adversarial Networks, which has been undergoing rapid development recently. In the first GAN [11], a model that can implicitly generate any probabilistic distribution was proposed. Then the conditional version of GAN [21] was proposed to enable generation under condition. Subsequent works include usage of convolution neural networks [28], and improve the training stability [30] followed by Wasserstein GAN [1] and Improved WGAN [12] which further made Generative Adversarial Networks reliable in real world applications. In our paper, we first train a conditional WGAN to generate single frame human pose, then we train another conditional sequence GAN to find the optimal latent code combination on the single pose generator to generate a pose sequence that look real and natural in order to perform video completion in the latent space of sequence generator.

**Optimization over Input Data** To specialize our general generative model to video prediction and completion, we model them as constrained video generation and adopt randomization of the input data to find the motion sequence that best matches the input frames. In the most recent work, back-propagation on input data is performed on image inpainting [43] to find the best match of corrupted image. Zhu et al [45] utilized such method to enable generative visual manipulation by optimizing on the latent manifold. Google DeepDream [22] also used back-propagation to generate dream-like images. Earlier, similar method has been employed to perform texture synthesis and style transfer [9, 10, 17]. In our paper, we similarly design specific loss and perform randomized optimization on the latent space.

**Skeleton to Image Transformation** Our two-stage model involves a second stage to transform human pose to pixel level image in order to generate video in pixel space, which has been attempted by various deep learning methods. Recent methods like [42, 19, 37] utilize GAN or multi-stage method to complete this task. We propose a simple yet effective supervised learning framework qualitatively comparable to state-of-the-arts.

## 3. Methodology

We present a general generative model that uniformly addresses video generation, prediction and completion problems for human motions. The model itself is originally designed for video generation, i.e., generating human action videos from random noise. We split the generation process into two stages: first, we generate human skeleton sequences from random noise, and then we transform from the skeleton images to the real pixel-level images. In Sec-tion 3.1 we will elaborate the model and methods we use to generate human skeleton motion sequences, and in Section 3.2 we will present our novel method for solving the skeleton-to-image transformation problem. Figure 2 illustrates our pipeline. Lastly, in Section 3.3, we will show that we can specialize this model without modification to accomplish video prediction and completion by regarding them as constrained video generation.

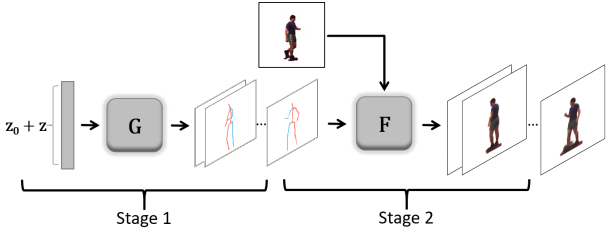

Figure 2. Overview of our two-stage video generation. In the first stage we generate skeleton motion sequences by $G$ from random noise, while in the second stage we use our skeleton-to-image transformer $F$ to transform skeleton sequence to image space.

## 3.1. General Generative Model

We propose a two-step generation model that generates human skeleton motion sequences from random noise.

Let $J$ be the number of joints of human skeleton in a video frame, and we represent each joint by its (x,y) location in image space. We formulate a skeleton motion sequence $V$ as a collection of human skeletons across $T$ consecutive frames in total, i.e., $V \in \mathbb{R}^{T \times 2J}$, where each skeleton frame $V_t \in \mathbb{R}^{2J}, t \in \{1 \cdots T\}$ is a vector containing all $(x, y)$ joint locations. Our goal is to learn a function $G : \mathbb{R}^n \rightarrow \mathbb{R}^{T \times 2J}$ which maps an $n$-dimensional noise vector to a joint location vector sequence.

To find this mapping, our experiments showed that human pose constraints are too complicated to be captured by an end-to-end model trained from direct GAN method [11]. Therefore, we switch to our novel two-step strategy, where we first train a *Single Pose Generator* $G_0 : \mathbb{R}^m \rightarrow \mathbb{R}^{2J}$ which maps a $m$-dimensional latent vector to a single-frame pose vector, and then train a *Pose Sequence Generator* $G_{PS} : \mathbb{R}^n \rightarrow \mathbb{R}^{T \times m}$ which maps the input random noise to the latent vector sequences, the latter of which can be transformed into human pose vector sequences through our *Single Pose Generator* in a frame-by-frame manner.

Figure 3 shows the overall pipeline and the results for each step in the procedure. The advantage of adopting this two-step method is that by training the single-frame generator, we enforce human pose constraints on each frame generated, which alleviate the difficulty compared to end-to-end GAN training thus enabling the model to generate longer sequences. Additionally, in order to generate different types of motions, we employ the Conditional GAN [21] method and concatenate an one-hot class vector indicating which class of motion to be produced to the input of our generators.

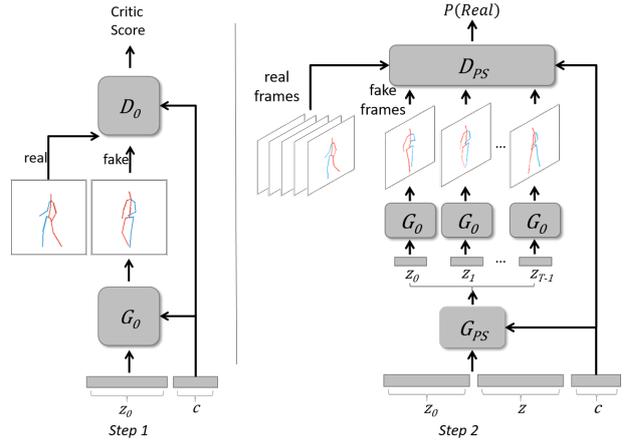

Figure 3. Illustration of our two-step generation pipeline. In step one (left) $G_0$ takes a random noise vector and outputs the generated pose vector. The $D_0$ then differentiate between real and fake pose vectors. Both inputs to $G_0$ and $D_0$ are concatenated with conditional class vector. In step two (right), $G_{PS}$ takes the random noise $z$ conditioned on the latent vector of the first frame and the class vector, and generates a sequence of latent vectors which can be transformed to pose vectors via $G_0$. Then $D_{PS}$ takes as input real/fake frames to determine $P(Real)$.

### 3.1.1 Single Pose Generator

In the first step, we employ the improved WGAN [12] method with gradient penalty for our adversarial training. We build a multilayer perceptron (MLP) for both our generator and critic with similar structures and add condition to the input of both of them according to Conditional GAN [21]. Our generator $G_0$ takes as input an $m$-dimensional latent vector $z_0$ concatenated with a one-hot class vector $c$ and outputs a pose vector $G_0(z_0|c)$. Our critic $D_0$ takes as input a real pose vector $x_0$ or a generated one, concatenated with $c$, yielding a critic score. The detailed architecture configurations are shown in Figure 4a, and are detailed in supplementary materials. Thus the WGAN objective is:

$$\min_{G_0} \max_{D_0 \in \mathcal{D}} \mathbb{E}_{c \sim p_c}[\mathbb{E}_{x_0 \sim p_{pose}}[D_0(x_0|c)] - \\ \mathbb{E}_{z_0 \sim p_{z_0}}[D_0(G_0(z_0|c)|c)]] \quad (1)$$

where $\mathcal{D}$ is the set of 1-Lipschitz functions, $p_c$ is the distribution of different classes, $p_{pose}$ is the distribution of the real pose data, and $p_{z_0}$ is the uniform noise distribution. Figure 5a shows some of our generated pose subject to given action class.

### 3.1.2 Pose Sequence Generator

In the second step, we use the normal GAN [11] method instead for training our *Pose Sequence Generator*, since in practice normal GAN performs better than WGAN for this specific task. The generator $G_{PS}$ generates a sequence of latent vectors, which are then fed into the *Single Pose Generator* resulting in a sequence of pose vectors $\hat{V}$, from a random noise vector $z$ conditioned on $z_0$ and $c$. Note that $z_0$ is a random noise vector describing the initial condition of the generated pose.

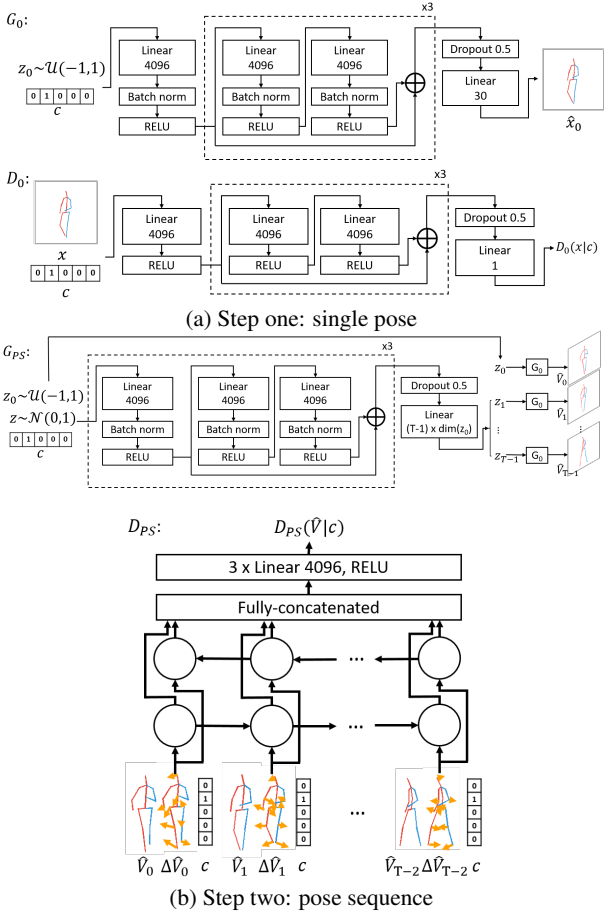

(a) Step one: single pose

(b) Step two: pose sequence

Figure 4. Two-step generation architecture. Detailed architecture configuration of step one and step two are shown in (a) and (b) respectively. Here ⊕ stands for element wise addition and ◯ stands for an LSTM cell.

In our implementation we generate latent vector sequences by generating the shifts between two consecutive frames, namely, the output of the network is $s_0, s_1, ..., s_{T-2}$ where $z_{t+1} = s_t + z_t$ for all $t \in \{0...T-2\}$ and $z_t$ is the latent vector for the $t$-th frame ($z_0$ is given from the noise distribution).

For the discriminator, we employ a bi-directional LSTM structure, whose input of each time step $t$ is the shift of consecutive frames $\Delta \hat{V}_t = \hat{V}_{t+1} - \hat{V}_t$ conditioned on $\hat{V}_t$ and $c$. The structural details are shown in Figure 4b. The objective function for the training in this step is:

$$\min_{G_{PS}} \max_{D_{PS}} \mathbb{E}_{c \sim p_c}[\mathbb{E}_{V \sim p_{video}}[\log D_{PS}(V|c)] + \\ \mathbb{E}_{z_0 \sim p_{z_0}, z \sim p_z}[\log(1 - D_{PS}(G_{PS}(z_0|z, c)|c))]] \quad (2)$$

where $\mathcal{P}_c$ is the distribution of different classes, $p_{video}$ is the distribution of the real video sequence data, $p_{z_0}$ is the uniform noise distribution and $p_z$ is the Gaussian noise distribution. In our implementation, we also add an L2 regularization term on the generated latent vector shifts for temporal smoothness, and thus we restrict the output latent

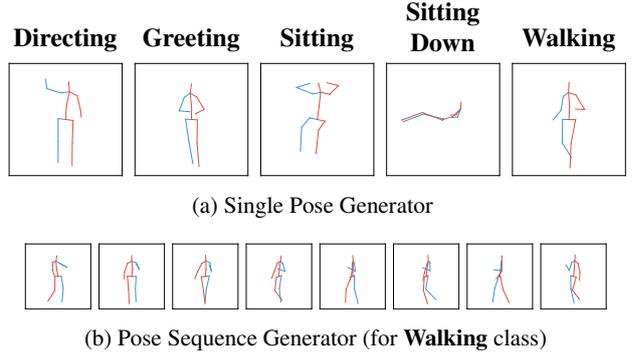

(a) Single Pose Generator

(b) Pose Sequence Generator (for **Walking** class)

Figure 5. Results demonstration for our two-step generation model. (a) Generated results for *Single Pose Generator* for five action classes. (b) Generated results for *Single Pose Generator*. Here we only show several examples for the walking class.

vectors in the range of $(-1, 1)$.

**Implementation Details** We use Adam Solver [16] at a learning rate of 0.001 for training *Single Pose Generator* and 5e-5 for training *Pose Sequence Generator*, both decaying by 0.5 after 30 epochs with $\beta_1$ being 0.5 and $\beta_2$ being 0.9. We set the weight of gradient penalty to be 10 and the weight of L2 regularization term for generated latent vector shift to be 0.1.

### 3.2. Skeleton to Image Transformation

In this stage, we train a skeleton-to-image transformation to convert pose space to image space. Formally, given an input pose vector $x \in \mathbb{R}^{2J}$ and a reference image $y_0 \in \mathbb{R}^{w \times h \times 3}$ where $h$ and $w$ are the width and height of images, we need to transform $\mathbf{x}$ to a pixel-level image $y \in \mathbb{R}^{w \times h \times 3}$. In order to make the dimensions of inputs well-aligned, we first convert the pose vector $x$ to a set of heat maps $S = (S_1, S_2, ..., S_J)$, where each heat map $S_j \in \mathbb{R}^{w \times h}, j \in \{1...J\}$ is a 2D representation of the probability that a particular joint occurs at each pixel location. Specifically, let $\mathbf{l_j} \in \mathbb{R}^2, (\mathbf{l_j} = (x_j, x_{j+1}))$ be the 2D position for joint $j$. The value at location $\mathbf{p} \in \mathbb{R}^2$ in the heat map $S_j$ is then defined as,

$$S_j(\mathbf{p}) = \exp(-\frac{\|\mathbf{p} - \mathbf{l_j}\|_2^2}{\sigma^2}) \quad (3)$$

where $\sigma$ controls the variance. Then our goal is to learn a function $F : \mathbb{R}^{w \times h \times J} \to \mathbb{R}^{w \times h \times 3}$ that transforms joint heat maps into pixel-level human images, conditioned on the input reference image. This function $F$ can thus be easily achieved by a feed-forward network. Note that we have the reference ground truth human images corresponding to each input poses in our training data, hence we can train this network in a supervised fashion.

**Skeleton-to-Image Network** To learn our function $F$, we employ a U-Net like network [29, 19] (i.e., convolutional autoencoder with skip connections as shown in Figure 6) that takes, as input, a set of joint heat maps $S$

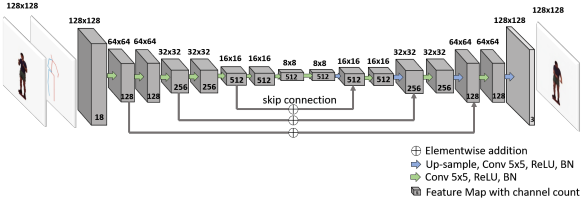

Figure 6. Skeleton to Image Network. Image sizes and feature dimensions are shown in the figure. Note that the input is of size $(128, 128, 18)$, which is the concatenation of $(128, 128, 18)$ reference image and $(128, 128, 15)$ heat maps for 15 joints

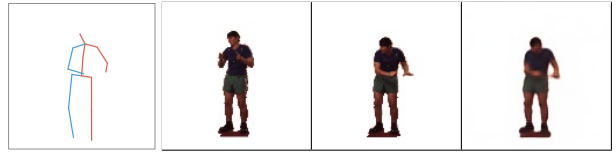

Figure 7. Illustration for skeleton-to-image training. From left to right: the input pose (encoded as heat maps), the input reference image, the corresponding ground truth and our results

and a reference image $y_0$ and produces, as output, a human image $\hat{y}$. For the encoder part, we employ a convolutional network which is adequately deep so that the final receptive field covers the entire image. For the decoder part, we use symmetric structure to gradually generate the image. To avoid inherit checkerboard artifact in transposed convolution layers, there has been several papers proposing solutions including sub-pixel convolution, resize and convolution etc [25, 31, 7]. In our case we apply nearest neighbor up-sampling followed by convolution layer in decoder.

**Loss Function** To train our skeleton-to-image network, we compare the output image with the corresponding reference ground truth image by binary cross entropy loss. We calculate the binary cross entropy loss for intensity values at each pixel, i.e.

$$\mathcal{L}_{bce} = -\frac{1}{k}\sum (1-y)\log(1-F(\mathbf{x}|y_0)) + y\log(F(\mathbf{x}|y_0)) \quad (4)$$

where $y$ is the reference ground truth image, $x$ is pixel and $k$ is the number of pixels. Our experiments show that only using binary cross entropy loss tends to produce blurry results. Hence, in order to enforce details in the produced images, we further employ a feature-matching loss (in some paper also referred as perceptual loss), as suggested in [4, 15]. We match the activations in a pre-trained visual perception network that is applied to both the reference ground truth image and the generated image. Different layers in the network represent different levels of abstraction, providing comprehensive guidance for our transformation function $F$ to generate more realistic images.

Specifically, let $\Phi$ be the visual perception network (We use VGG-19 [33]), and $\Phi_l$ be the activations in the $l$-th layer. Our feature-matching loss is defined as,

$$\mathcal{L}_2 = \sum_l \lambda_l \|\Phi_l(F(\mathbf{x}|y_0)) - \Phi_l(y)\|_1 \quad (5)$$

where $\lambda_l$ is the weight for the $l$-th layer, which are manually set to balance the contribution of each term. For layers $\Phi_l$, we use 'conv1_2', 'conv2_2', 'conv3_2', 'conv4_2' and 'conv5_2' in VGG-19 [33].

The overall loss function for our skeleton-to-image network is therefore defined as

$$\mathcal{L} = \mathcal{L}_1 + \lambda \mathcal{L}_2 \quad (6)$$

where $\lambda$ denotes the regularization factor of feature matching loss.

**Implementation Details** We train our network with Adam Solver [16] at a learning rate of 0.001 and $\beta_1$ of 0.9. For the feature matching loss we set $\lambda = 0.01$. Our architecture details are shown in Figure 6. In the encoder we have 8 convolution layers with $5 \times 5$ filter size, alternating stride of $2, 1, 2, 1, 2, 1, 2, 1$ and "same" padding. In the decoder we have 3 $(upsample, conv, conv)$ modules followed by 1 $(upsample, conv)$ module, where $upsample$ stands for nearest neighbor resize with $scale\,factor = 2$ and $conv$ stands for convolution layer with $5 \times 5$ filter size and 1 stride size with "same" padding.

### 3.3. Prediction and Completion

To uniformly address video completion and video prediction, we model them as constrained video generation, which is ready to be defined by the general generative model. We optimize on the latent space in order to achieve our goal. For simplicity, the optimization is conducted based on generated pose sequence, and we can transform to complete video by our skeleton-to-image transformer using the completed pose sequence. We utilize state-of-the-art human pose estimation methods like [23] to obtain pose sequences from videos.

#### 3.3.1 Video Completion

To fill in missing frames of a video, our method utilizes the generator $G$ trained with full-length human pose sequence. We assume that the learned encoding manifold $\mathbf{z}$ is effective in representing $p_{data}$. We aim to perform video completion by finding the encoding $\hat{\mathbf{z}}$ on the manifold that best fits the input frames constraint. As illustrated in Figure 8, we can generate the missing content by using the trained generative model $G$.

**Objective Function** We regard the constrained video generation problem as an optimization problem. Let $I \in \mathbb{R}^{t \times 2J}$ be the input frames which acts as constraint and $\mathbf{z}$ denote the learned encoding manifold of $G$. With these notations, we define the optimal completion encoding $\hat{\mathbf{z}}$ by:

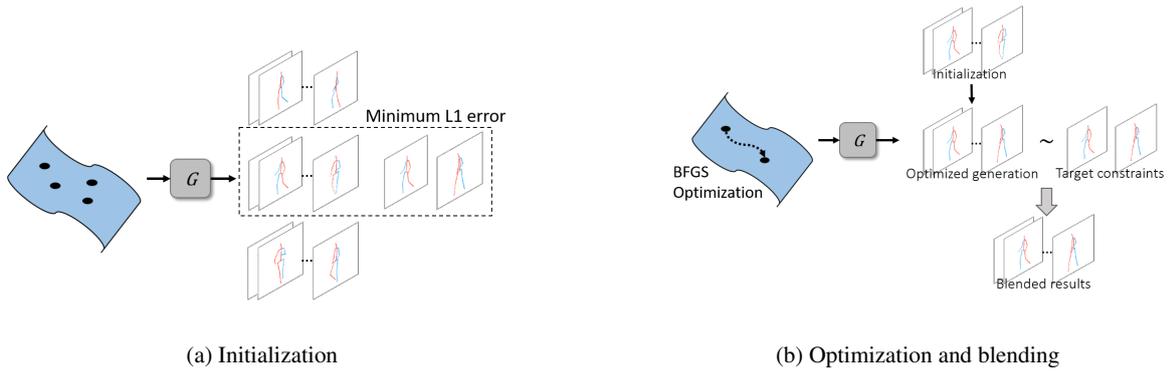

Figure 8. Our completion/prediction pipeline. (a) Initialization: we randomly sample from the latent space and compare L1 error with the constraint frames. Dashed box shows the best initialization chosen. (b) We run randomized optimization algorithms starting at our initialization. We blend the constraints and the generated results as our final output.

$$\hat{\mathbf{z}} = \arg\min_{\mathbf{z}}\{\mathcal{L}_c(\mathbf{z}|\mathbf{I}) + \alpha \times \mathcal{L}_p(\mathbf{z})\}, \quad (7)$$

where $\mathcal{L}_c$ denotes the contextual L1 loss between the constrained frames and corresponding generated frames and $\mathcal{L}_p$ denotes the perceptual loss of generated frames, i.e. "realness" of the pose sequence. $\alpha$ denotes a regularization factor of the perceptual loss. $\mathcal{L}_c$ and $\mathcal{L}_p$ are defined as follows:

$$\mathcal{L}_c(\mathbf{z}|\mathbf{I}) = \sum_{i \in I} |G(\mathbf{z})_i - \mathbf{I}_i| \quad (8)$$

$$\mathcal{L}_p(\mathbf{z}) = -\log(D(G(\mathbf{z}))) \quad (9)$$

where $\mathbf{I}$ is the set of constrained frames and $z$ is some latent vector, $i$ denotes the index of frames in $\mathbf{I}$; $i$ can be arbitrary numbers subject to the given constraints. By optimizing Eq. (7), we obtain a full generated sequence $G(\hat{\mathbf{z}}) \in \mathbb{R}^{T \times 2J}$ which is the "closest" match to the input frames.

**Two-Step Optimization** In order to optimize Eq. (7), we employ a two-step method illustrated in Figure 8a.

To address the optimization of such highly non-convex latent space, we first randomly sample from the latent space and compare the loss of Eq. (7) to find the best initialization, namely $\mathbf{z}_0$.

As proposed in [45], taken the optimal initialization $\mathbf{z}_0$ as the starting point, we apply Limited Broyden-Fletcher-Goldfarb-Shanno optimization (L-BFGS-B) [2], a quasi-newton optimization algorithm well known for its optimality, on the $(n+m)$-dimension latent space in order to find the optimal completion result, namely $\hat{\mathbf{z}}$.

**Video Blending** After generating $G(\hat{\mathbf{z}})$ in pose space, the fully completed video can be simply obtained by stacking the input frames with the generated frames. However, slight shift and distortion in motion are observed as our method may not always produce perfect alignment with the input. To address this, we use Poisson blending [27] to smooth our final pose sequence to make them more natural while satisfying the input constraints. The key idea is to maintain the gradients on the temporal direction of $G(\hat{\mathbf{z}})$ to preserve motion smoothness while shifting the generated frames to match the input constraint. Our final solution, $\hat{\mathbf{x}}$, can be obtained by

$$\hat{\mathbf{x}} = \arg\min_{\mathbf{x}} \|\nabla_t \mathbf{x} - \nabla_t G(\hat{\mathbf{z}})\|_2^2,$$
$$\text{s.t. } \mathbf{x}_i = \mathbf{I}_i \text{ for } i \in \mathbb{R}^{t \times 2J} \quad (10)$$

where $\nabla_t$ is the gradient operator on the temporal dimension. The minimization problem contains a quadratic term, which has a unique solution [27]. We will show qualitative results that the proposed blending preserves the naturalness of the videos while better aligning with the input frame constraints.

### 3.3.2 Video Prediction

Video prediction can be solved under the same general framework as it can be essentially interpreted as video generation with first few frames as constraint. A figure of illustration is in Figure 8, where we exemplify how to perform future predictions by using the trained generative model $G$.

Formally, let $I \in \mathbb{R}^{t \times 2J}$ be consecutive frames at time step $0$ to $t$ as input, we generate future frames $G_t, G_{t+1}, \cdots G_T$ so that $I_0, I_1, \cdots, I_t, G_{t+1}, \cdots G_T$ form a natural and semantically meaningful video. To achieve such goal, we model video prediction as video generation with first few frames as constraint. In other words, we perform the same steps in 3.3.1 with the input described above, then we can obtain a completed video sequence where $t+1$ to $T$ frames are generated future frames.

## 4. Experiments

### 4.1. Dataset

We evaluate our model on Human3.6m dataset [13]. This dataset consists of 896 human motion videos captured from 7 different subjects performing various classes of motion (e.g., walking, sitting and greeting, etc.). The videos were captured at high-resolution 1000×1000, at frame rate of 50 fps. The dataset provides reference ground truth 2D human poses (joint locations).

In our experiments, in order to reduce redundant frames and encourage larger motion variations, we subsample the video frames to 16 fps. The action classes we select are 'Direction', 'Greeting', 'Sitting', 'Sitting Down', 'Walking', all of which contain large human motions.

For our skeleton sequence generation task, we randomly select 5 subjects as training set and reserve 2 subjects as testing set. We normalize the ground truth 2D pose annotations so that the hip joints are all centered at the middle of the frames and the limb lengths are of the same scale. For our skeleton-to-image transformation task, we treat the unchosen action classes as training set, and our chosen 5 action classes as testing set. Human images are extracted from the original video frames using a 512×512 window, and then are resized into 128×128. Note that, since our major concern is human motion, we thus subtract all the backgrounds and generate the foreground human figure only. Background completion can be achieved by some traditional methods as in [6, 26].

### 4.2. Evaluation

Evaluating the quality of synthesized videos is a difficult problem in the case of video generation. Traditional methods such as per-pixel mean-squared error does not apply here as there can be multiple visually plausible solutions. Furthermore, the pixel based MSE does not measure the temporal smoothness and human-likeness which we aim to model in this paper.

In order to evaluate the visual quality of our results, we measure whether our generated videos are adequately realistic such that a pre-trained recognition network can recognize the object and action in the generated video. This method is inherently similar to the Inception Score in [37, 30], object detection evaluation in [39] and Semantic Interpretability in [44]. Two-stream model is first proposed by Yan et al [32] in 2014 and further improved by Wang et al [38]. Thus we fine-tune a state-of-the-art two-stream video action recognition [38] network on our dataset. Then we evaluate the following two scores measuring the visual quality of generated image frames and video sequences respectively:

**Inception Score for frames** One criterion regarding evaluating scores for video is that they should reflect if the video contains natural images along the sequence. Thus we calculate the inception score [30] based on the output classification result of the RGB stream [38] for each frame generated as the evaluation metric. The average score across the whole video should reflect the overall image quality. Additionally, we also show the Inception Score obtained at each time step, which gives us a detailed snapshot of how the quality of video vary over time.

**Inception Score for videos** As proposed in [37], we evaluate the inception score [30] based on the fused classification results from our two-stream action classifier. By taking in to consideration the motion flow across the whole video, the output classes serve as an accurate indicator of the actions perceived in the video. Thus such score can give an overall quality of the full video sequence.

### 4.3. Baselines

We present several baseline methods to provide comparisons of our results with results from previous methods.

For Video Generation, our baseline is *Video-GAN* (VGAN) [35]. This approach trains a GAN that generates videos in pixel space. It is first successful attempt on video generation with deep learning methods.

For Video Prediction, the first baseline is *PredNet*, a predictive network [18]. This approach is one of the latest results in video prediction. The second baseline is a *Multi-Scale GAN* (MS-GAN) in pixel space proposed by Mathieu et al [20]. This approach has been quite successful in various video prediction tasks including both human action videos. The third baseline is *PoseVAE*, a sequential model proposed in [37], which utilized pose representation and have produced state-of-the-art results.

For Video Completion, our baseline is *Conditional Video-GAN* (cond-VGAN) [35]. The model is capable of predicting next frames given input as shown in the paper, therefore we adopt it to video completion by changing its input to the first and last frame.

## 5. Results

For video generation, we generate videos from random noise vectors with dimensions consistent with the proposed models. For video prediction, we feed the first 4 frames as inputs, i.e. the baselines make prediction based on the input 4 frames, and our model generates videos with the first 4 frames as constraints. For video completion, we fix the the first and the last frames as constraints. In order to calculate the proposed metrics, we randomly generate 320 50-frame video samples for each method (except for the Video-GAN method [35] which is fixed by architecture to generate only 32 frames). For prediction results, 4 preceding frames are fed as input. For completion results, the first and the last frames are fed as input.

### 5.1. Qualitative Results

In Figure 9 we show the qualitative results of our model, in comparison with other state-of-the-art video generation and prediction methods. Since the results are videos, we strongly suggest readers to check our supplementary materials, which provide better visual comparisons than the sample frames shown in the paper. The baseline methods are all fine-tuned/re-trained on our Human3.6m dataset [13]. We show generated results for each of our selected classes. Due to the page limit, we only show the beginning and the middle frames in the result videos.

By examining the results, we find that our model is capable of generating plausible human motion videos with high visual quality. By examining the image quality, we find that our model generates the most compelling human images, while other models tend to generate noisy (particularly Video-GAN) and blurry results due to their structural limitations. By examining the video sequences (provided in supplementary materials), we find that our model can generate natural and interpretable human motions. A key distinction here is that we are able to produce large-scale and detailed motion. For instance, the walking example shows clearly how the person moves his hands and legs in order to complete this action. Another important observation is that, our results maintain high quality over the entire time interval, while the others' quality (especially prediction models) tend to drop significantly after first few predictions. In the baseline prediction models, the human subjects are fading over time.

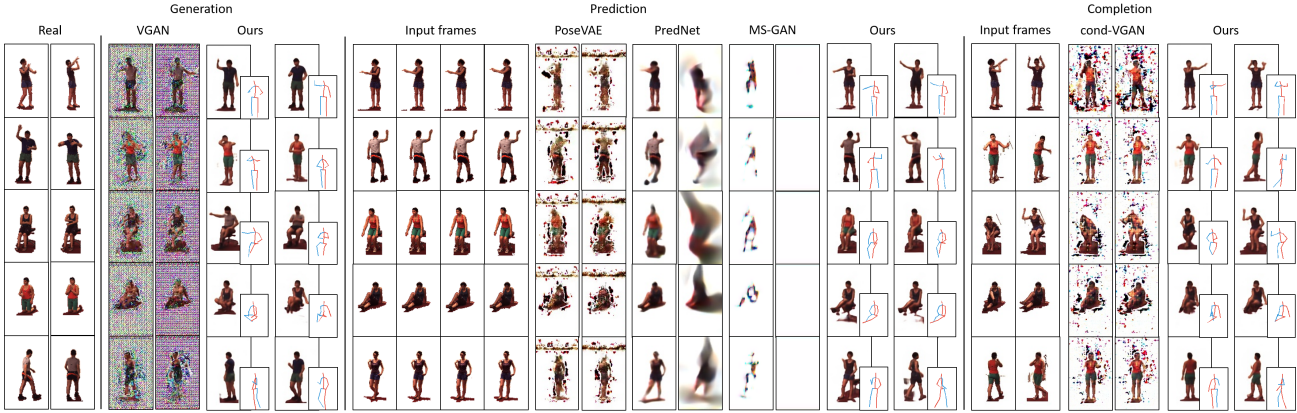

Figure 9. Qualitative comparisons. Each image-pair column corresponds to a generation method (the first column is real data), and columns are grouped together in the order of generation, prediction and completion, respectively. Each row corresponds to an action class, from top to bottom: Direction, Greeting, Sitting, Sitting Down, Walking. For each method we show the 10th and the 40th frames. For our method we also show the generated pose results.

Table 1. Frame and Video Inception Score (IS)

| Method | frame-IS | video-IS |
|---|---|---|
| Real | $4.53 \pm 0.01$ | $4.63 \pm 0.09$ |
| VGAN[35] | $1.53 \pm 0.04$ | $1.40 \pm 0.16$ |
| Ours | $\mathbf{3.99 \pm 0.02}$ | $\mathbf{3.99 \pm 0.18}$ |
| PoseVAE[37] | $1.91 \pm 0.01$ | $2.17 \pm 0.11$ |
| PredNet[18] | $2.60 \pm 0.04$ | $2.94 \pm 0.15$ |
| MS-GAN[20] | $1.48 \pm 0.01$ | $1.88 \pm 0.10$ |
| Ours | $\mathbf{3.87 \pm 0.02}$ | $\mathbf{4.09 \pm 0.15}$ |
| cond-VGAN | $2.35 \pm 0.02$ | $2.00 \pm 0.06$ |
| Ours | $\mathbf{3.91 \pm 0.02}$ | $\mathbf{4.10 \pm 0.07}$ |

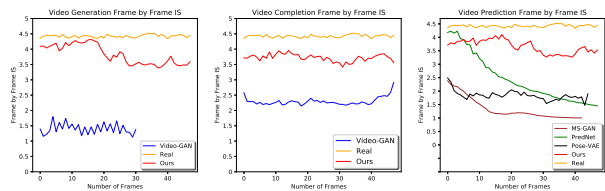

Figure 10. Generation, Completion and Prediction.

### 5.2. Quantitative Results

Table 1 tabulates our quantitative evaluation results, "frame-IS" stands for Inception Score for frames, and "video-IS" stands for Inception Score for videos. While the ground truth (real) videos have the largest Inception Scores of both types, which coincides with our intuition, our generated videos have the highest scores among all the competing methods. This suggests that our model generates videos that have meaningful visual features in both image and video (spatial and temporal) domains closer to real videos, thus further indicating that our videos are more realistic. We also observe that other methods have much lower scores than ours, and VGAN and [20] are even worse than PredNet. All the statistics are consistent with our qualitative results, which have demonstrated great discrepancy in visual perception.

Figure 10 shows a comparison of frame-by-frame Inception Score. We find that the ground truth videos maintain the highest scores at all time steps, and our results have considerably high scores closest to the ground truth quality. A more important observation is that, for the compared prediction models, PredNet [18] and [20], the scores tend to fall with time, indicating that the image quality is deteriorating over time in the long run. Although PoseVAE [37] does not decline, its overall image quality is much lower than ours. This observation is consistent with our qualitative evaluation. An explanation might be the model's reliance on the input preceding frames, whose quality gradually drops since they are generated images. For completion baseline, the score tends to be higher at both ends, due to the restriction on the first and last frame.

## 6. Conclusion and Future Work

We present a general generative model that addresses the problem of video generation, video completion and video prediction uniformly. By utilizing human pose as intermediate representation with our novel two-step generation strategy, we are able to generate large-scale human motion videos with longer duration from scratch. We are then able to solve the later two problems by constrained generation using our model. We find that our model is able to generated plausible human action videos both from scratch and under constraint, which surpasses current methods both quantitatively and visually.

Future directions include: in terms of improvement, more data can be incorporated in training both the motion generator and the skeleton-to-image transformer to increase the robustness and generality of our model. Multi-scale progressive training in second stage may also further increase the sharpness of the generated images. In terms of general video generation, it is possible to incorporate RNN to generate variable length videos.